\def\year{2022}\relax
\documentclass[letterpaper]{article} % DO NOT CHANGE THIS
\usepackage{aaai22}  % DO NOT CHANGE THIS
\usepackage{times}  % DO NOT CHANGE THIS
\usepackage{helvet}  % DO NOT CHANGE THIS
\usepackage{courier}  % DO NOT CHANGE THIS
\usepackage[hyphens]{url}  % DO NOT CHANGE THIS
\usepackage{graphicx} % DO NOT CHANGE THIS
\usepackage{natbib}  % DO NOT CHANGE THIS AND DO NOT ADD ANY OPTIONS TO IT
\usepackage{caption} % DO NOT CHANGE THIS AND DO NOT ADD ANY OPTIONS TO IT
\DeclareCaptionStyle{ruled}{labelfont=normalfont,labelsep=colon,strut=off} % DO NOT CHANGE THIS
\usepackage{amsmath}
\usepackage[ruled,linesnumbered]{algorithm2e}
\usepackage{subfigure} 

\usepackage{newfloat}
\usepackage{listings}
 \usepackage{hyperref} %-- This package is specifically forbidden

\title{Juvenile state hypothesis: What we can learn from lottery ticket hypothesis researches?}
\author{
Di Zhang
}
\affiliations{
University of Science and Technology of China\\
trotsky@mail.ustc.edu.cn

}

\usepackage{bibentry}
\begin{document}

\maketitle

\begin{abstract}
    \begin{quote}
    The proposition of lottery ticket hypothesis revealed the relationship between network structure and initialization parameters and the learning potential of neural networks. The original lottery ticket hypothesis performs pruning and weight resetting after training convergence, exposing it to the problem of forgotten learning knowledge and potential high cost of training. Therefore, we propose a strategy that combines the idea of neural network structure search with a pruning algorithm to alleviate this problem. This algorithm searches and extends the network structure on existing winning ticket sub-network to producing new winning ticket recursively. This allows the training and pruning process to continue without compromising performance. A new winning ticket sub-network with deeper network structure, better generalization ability and better test performance can be obtained in this recursive manner. This method can solve: the difficulty of training or performance degradation of the sub-networks after pruning, the forgetting of the weights of the original lottery ticket hypothesis and the difficulty of generating winning ticket sub-network when the final network structure is not given. We validate this strategy on the MNIST and CIFAR-10 datasets. And after relating it to similar biological phenomena and relevant lottery ticket hypothesis studies in recent years, we will further propose a new hypothesis to discuss which factors that can keep a network juvenile, i.e., those possible factors that influence the learning potential or generalization performance of a neural network during training.
    \end{quote}
\end{abstract}
\section{Introduction}
After the proposition of the lottery ticket hypothesis \citep{frankle_lottery_2019}, more effective methods for neural network pruning appeared, and at the same time, new methods for neural network architect searching(NAS) were emerging. NAS algorithm searching for the network's architectural parameters to construct the mapping function from the network's input to the network's output \citep{ren2021comprehensive}. Network pruning reduces the total number of the network's parameters by removing those parameters of lesser importance from the network weight parameters \citep{lecun_optimal_1990}. The main difference between NAS and network printing is that the former is based on the network's architecture, while the latter is based on the network's parameters. To some extent,they applied similar procedures to neural networks since network pruning can cut unnecessary links between neurons by eliminating the network's unimportant parameters and finding a better architecture for the network.

Therefore, we can consider the pruning algorithm as a network structure optimization operator. The pruning algorithm optimizes the inter-neuron connection structure at a finer granularity when the neural architecture search algorithm for the network layer width and the path of forwarding propagation flow.

However, in practice, pruning algorithms are usually used to prune neural networks trained close to convergence to optimize their inference speed and memory efficiency \citep{LIANG2021370}, and many researchers point out that continuing training on the neural network after pruning could cause performance degradation on the validation set, and that training from scratch is hard using pruned structures \citep{DBLP:journals/corr/LiKDSG16,DBLP:journals/corr/HanPTD15}. To address the latter issue, Frankle et al. proposed the lottery ticket hypothesis(LTH) that the initialization is a key point for one neural network and its sub-networks to get better performance on the validation set during a similar training process. By pruning the neural networks close to the training fit, the unmasked network weights are re-trained after re-adopting the parameters from the original initialization, and the process is repeated until obtaining a sub-network with the required amount of parameters and accuracy.

Even so, in large neural networks, the overhead of the pruning-reset-retraining cycle is unaffordable, and Frankle et al. then tried using weights after trained for iterations to reset the pruned network \citep{frankle2019stabilizing,renda2020comparing} instead of the totally original initial weights. But the recovery of the original parameters and weights means forgetting of most knowledge learned in training, even some researcher assumed that pruned architecture keeps the memory of the network's inductive bias \citep{frankle_lottery_2019,cohen_inductive_2016}.

But after trying to add new trainable layers and parameters to pruned near-convergent neural networks,also known as \emph{winning tickets}, we found that the tested network could continue training until convergence nearly at a higher test accuracy.

Therefore, we proposed a new iterative pruning algorithm that is based on a generalized inference of LTH, This pruning algorithm combines a simple network architecture search Process and pruning-retraining cycle into a pruning-growing-retraining approach to shape a deep and thin neural network to meet the requirement of parameter's amount and test accuracy.

Then we further generalize the lottery ticket hypothesis into recursive lottery ticket hypothesis, and by this hypothesis and other inferences of lottery ticket hypotheses, we will further explore the juvenile state of neural networks beyond the luckiness of initialization to try to explain what determines the learning potential and convergence speed of neural networks.

\section{Approach}
\subsection{Lottery Ticket Hypothesis}
First, we would have a review of the naive Lottery ticket hypothesis (LTH).

\subsubsection{Notion} We follow the description of the lottery ticket hypothesis by Frankle et al. \citep{frankle_lottery_2019}. but with an additional symbol $\mathbf{\theta}$ to represent the architecture of the network and shaping the vector space of weight parameters. Given a feed-forward neural network $f$ with architecture $\mathbf{\theta}$ , weights $\mathbf{\omega} $ and taking $x$ as input.

Furthermore,the weight parameter $\mathbf{\omega}$ defined a set of parameters $\mathbf{\omega}_i$ for each layer $i$. The architecture $\mathbf{\theta}$ defines the number of neurons, width $w$ in each layer $i$ and the number of layers $L$, as well as the path of forwarding propagation flow especially inter-neuron connections through mask matrix $m_i$ in each layer $i$.

\subsubsection{Formal Definitions}At start, the network weight parameters $\mathbf{\omega}$ are randomly initialized with $\mathbf{\omega}^0 \sim \mathcal{D}_{\mathbf{\omega}}$to adapt the structure $\mathbf{\theta}^0$ of the network. Then,the network $f(x;\mathbf{\theta}^0,\mathbf{\omega}^0 )$ trained on a dataset $\mathbf{D}$ and optimized by a gradient descent algorithm optimizer $\mathcal{L} $ close to convergence.

Then the network  $f(x;\mathbf{\theta}^0,\mathbf{\omega}^1 )$is pruned by removing the parameters of lesser importance from the network weight parameters $\mathbf{\omega}$,and produce mask matrices $m_i$ for each layer $i$ to define inter-neuron connections between layers.

The cycle of training and pruning could execute iteratively until the final network $f(x;\mathbf{\theta}^*,\mathbf{\omega}^* )$ achieve the expected test accuracy $Acc^*$ and compression ratio $r_{comp}^*$of the number of parameters or meet the maximum iterations $j$, and produce a mask matrix $m$.

The network $f(x;\mathbf{\theta}^*,\mathbf{\omega}^* )$is said to be a \emph{lottery ticket} network if the following conditions are satisfied:

1.$\frac{\left\| \mathbf{\omega}^*  \right\|_0}{\left\| \mathbf{\omega}^0 \right\|_0}  \le r_{comp}^*$

2. Existing a $(\mathbf{\theta}',\mathbf{\omega}')$ with the same architecture of $\mathbf{\theta}^*$ and the same number of parameters of $\mathbf{\omega}^*$ and a policy $p$, follow this policy $p$,we can training network$f(x;\mathbf{\theta}',\mathbf{\omega}' )$ close to convergence,and achieve a new test accuracy $Acc'$ close enough to $Acc^*$.

So,We take the identifying method which frankle et al. used in their work \citep{frankle_lottery_2019} as the weight reset policy,its work process can be described by the following pseudo-code:

\begin{algorithm}
    \caption{Weight reset policy}
    \SetAlgoLined
    \textbf{Given} $f(x;\mathbf{\theta}^0,\mathbf{\omega}^0 )$ \\
    \KwResult{\emph{winning ticket} $f(x;\mathbf{\theta}',\mathbf{\omega}' )$  }
    initialize $\mathbf{\omega}^0 \sim \mathcal{D}_{\mathbf{\omega}}$to adapt the structure $\mathbf{\theta}^0$ \\
    record $(\mathbf{\theta}^0,\mathbf{\omega}^0)$ \\
    training and pruning $f(x;\mathbf{\theta}^0,\mathbf{\omega}^0 )$ for $j$ iterations,arriving at $f(x;\mathbf{\theta}^*,\mathbf{\omega}^* )$\\
    $\omega' \leftarrow m \odot \omega^0 $, reset $\mathbf{\omega}^*$ to $\mathbf{\omega}^0$ according to the mask matrix $m$,and denote as  $\mathbf{\omega}'$\\
    produce \emph{winning ticket} $f(x;\mathbf{\theta}',\mathbf{\omega}' )$\\
    \textbf{(Optional)} Verifying the \emph{winning ticket} $f(x;\mathbf{\theta}',\mathbf{\omega}' )$\\
    training $f(x;\mathbf{\theta}',\mathbf{\omega}' )$ for $j$ iterations and get a new test accuracy $Acc'$\\
    \textbf{If}{$\frac{\left\| \mathbf{\omega}'  \right\|}{\left\| \mathbf{\omega}^0 \right\|}  \le r_{comp}^*$ \textbf{And} $Acc'$ is closing enough to $Acc^*$}\\
    \textbf{Then}{$f(x;\mathbf{\theta}',\mathbf{\omega}' )$ is a \emph{winning ticket}}\\
    \textbf{End}
\end{algorithm}

With this policy, we can solve the accuracy degradation problem when we continue to train the pruned neural network.

That is, by resetting the network weights to the original initialization values, we can retrain the network with a smaller number of parameters and keep the final test accuracy with a little degradation or even better compared to the network with a larger number of parameters.
\subsection{Recursive Lottery Ticket Hypothesis}
Then,we propose the recursive lottery ticket hypothesis (RLTH), which is a generalization of the former naive lottery ticket hypothesis allows us to obtain a new \emph{winning ticket} with more trainable parameters from an existing \emph{winning ticket} by a recursive way like pruning-growing-retraining cycle.

Up to this point, we have used the pruning algorithm $\mathcal{L}_{prun}$ only for optimizing the number of parameters of a trained neural network with a fixed structure $\theta$.

In the following part, we will go a step further and use the pruning algorithm $\mathcal{L}_{prun}$ as a fine-grained structure optimization operator to get a sub-network with optimal structure from original network. And try to extend the existed  \emph{winning ticket} with new neural network layers and trainable parameters, and obtain a new \emph{winning ticket} recursively by a similar training-pruning process.

Under the weight reset strategy, the network weights must be reset if we want continue training on the pruned structure without compromising the test accuracy, but it implies discarding and forgetting the learned knowledge while preserving only a little learned inductive bias preserved by the structural information.

Assuming that we have already obtained a \emph{winning ticket} sub-network by the original lottery ticket hypothesis, we can make following assumptions to induce the recursive lottery ticket hypothesis, which is a generalized inference of original lottery ticket hypothesis.

\textbf{Inference 1}: When neural network $f(x;\mathbf{\theta}^*,\mathbf{\omega}^* )$ is a \emph{winning ticket} obtained following a policy  $p$ from neural network $f(x;\mathbf{\theta}^0,\mathbf{\omega}^0 )$, and $m$ is a mask matrix, a \emph{winning ticket} structure parameterized by $\theta^*$ is a sub-network  of  structure parameterized by $\theta^0$,and on their corresponding graphical representations $G^*,G^0$, Also meet that $G^*$ is a \emph{spanning subgraph} of $G^0$.

Then we can construct a new father graph of that \emph{winning ticket} by add new nodes and links to existing graph through specific network layers with new trainable parameters.

\textbf{Inference 2}: Given a  \emph{winning ticket} network$f(x;\mathbf{\theta}^*,\mathbf{\omega}^* )$  ,there existing a \textbf{specific} network which structure parameterized by $\theta'$ with weights $\omega'$.

by adding that,we can construct a new neural network $f(x;\mathbf{\theta}^1,\mathbf{\omega}^1 )$, where
$$
\left\{\begin{aligned}
\theta^{1} &=\theta^{*} \cup \theta^{\prime} \\
\omega^{1} &=\omega^{*} \cup \omega^{\prime}
\end{aligned}\right.
$$
and according to the original LTH, neural network $f(x;\mathbf{\theta}^*,\mathbf{\omega}^* )$ can also be a \emph{winning ticket} sub-network of $f(x;\mathbf{\theta}^1,\mathbf{\omega}^1 )$,only if there exisit a policy $p$ and a pruning algorithm $\mathcal{L}$ could train and prune $f(x;\mathbf{\theta}^1,\mathbf{\omega}^1 )$ into $f(x;\mathbf{\theta}^*,\mathbf{\omega}^* )$.

Further more, we can also learn a simple fact from original LTH, that one neural network could have more than one \emph{winning ticket} sub-networks. Since our goal is to get a new \emph{winning ticket} sub-network,if our training policy and pruning algorithm meet the requirements fortunately, we can relax the premise of "specific extending layer" of Inference 2. Then the old \emph{winning ticket} sub-network can be extended with a new randomly initialized layer, and thus obtain a new \emph{winning ticket} through above training-pruning cycle.

So in summary, we can draw a generalized conclusion from the original lottery ticket hypothesis.

\textbf{inference 3} Given a \emph{winning ticket} neural networks $f(x;\mathbf{\theta}^*,\mathbf{\omega}^* )$ and a randomly initialized extending layer which struture parameterized by $\theta'$ with weights $\omega'$,  there exisit a policy $p$ and a pruning algorithm $\mathcal{L}$ could train and prune its extended network $f(x;\mathbf{\theta}^1,\mathbf{\omega}^1 )$ and produce a new \emph{winning ticket} sub-network.  

That is the Recurisive lottery ticket hypothesis(RLTH). With this conclusion we can start training with a simple small neural network and expand it after getting a \emph{winning ticket} sub-network. And repeat for several times, eventually getting a deeper \emph{winning ticket} recursively without forgetting of learned knowledge during the training process.

Formally, the basic training policy for the recursive lottery ticket hypothesis can be outlined in the following form.

\begin{algorithm}
    \caption{Basic structure growing policy}
    \SetAlgoLined
    \textbf{Given} $f(x;\mathbf{\theta}^0,\mathbf{\omega}^0 )$ \\
    \KwResult{\emph{winning ticket} $f(x;\mathbf{\theta}',\mathbf{\omega}' )$  }
    initialize $\mathbf{\omega}^0 \sim \mathcal{D}_{\mathbf{\omega}}$to adapt the structure $\mathbf{\theta}^0$ \\
    \While{sub-network can't meet requirements}
    {
    	training and pruning $f(x;\mathbf{\theta}^0,\mathbf{\omega}^0 )$ for $j$ iterations,arriving at $f(x;\mathbf{\theta}^*,\mathbf{\omega}^* )$\\
    	initialize a new layer $\mathbf{\omega}^1 \sim \mathcal{D}_{\mathbf{\omega}}$to adapt the structure $\mathbf{\theta}^1$ \\
        $\theta^{1} \leftarrow\theta^{*} \cup \theta^{\prime} ,\omega^{1} \leftarrow\omega^{*} \cup \omega^{\prime}$ \\
        construct a new sub-network $f(x;\mathbf{\theta}',\mathbf{\omega}')$ \\
    }
    produce \emph{winning ticket}$f(x;\mathbf{\theta}',\mathbf{\omega}')$\\
    \textbf{(Optional)} Verifying the \emph{winning ticket} $f(x;\mathbf{\theta}',\mathbf{\omega}' )$\\
    training $f(x;\mathbf{\theta}',\mathbf{\omega}' )$ for $j$ iterations and get a new test accuracy $Acc'$\\
    \textbf{If}{$\frac{\left\| \mathbf{\omega}'  \right\|}{\left\| \mathbf{\omega}^0 \right\|}  \le r_{comp}^*$ \textbf{And} $Acc'$ is closing enough to $Acc^*$}\\
    \textbf{Then}{$f(x;\mathbf{\theta}',\mathbf{\omega}' )$ is a \emph{winning ticket}}\\
    \textbf{End}
\end{algorithm}

This approach allows the neural network to grow like an oyster, growing to a deeper structure and keeping the structurally optimal with a proper parameters amount at the same time.

\section{Experiment}
\subsection{RLTH on Dense Networks}
First, we test the weight reset policy of the original LTH and the basic structure growing policy of RLTH on the MNIST dataset \citep{lecun_gradient-based_1998}.The experiment was executed on Google Colab platform. 

Then we iteratively training neural networks for 5 times to the same parameter amount by former two policies without any hyperparameter tunings, this process is a circle of training, obtain \emph{winning ticket} and training again.

The neural network to be trained with weight reset policy iteratively is a random-initialized dense network. The network have a input layer project the input image to a vector of 128 elements. And those 5 hidden layers have been set to a linear layer with a shape of (128,128) and a ReLU activation layer. Finally, the output layer produce the category probability of the output image with a Softmax activation layer.

And The neural network to be trained with structure growing policy have the same input and output layer and have no hidden layers at begin. When iteratively training, hidden layers would be insert to the network via so-called structure growing operation. To simplify the structure growing operation, we insert a linear layer with a shape of (128,128) and a ReLU activation layer per iteration. And to avoid the interruption of the training process, the inserted layer was initialized with a identity matrix for weight and a zero matrix for bias.

Notion that, in more complicated scenarios, the structure growing operation can be implement more properly by a sophisticated methods like other NAS algorithms.

For the convenience of reproduction, The other settings of the experiment are as follows. All random seeds set to 42. Optimizer of networks was Adam \citep{kingma2017adam} with a initial learning rate of 0.001. And the loss function is using categorical cross entropy. The prunning algorithm is Level pruner with a sparsity ratio of 95\% from NNI toolbox \citep{noauthor_microsoftnni_2021}.

\begin{figure}[htbb]
\subfigure[training]{
\label{Fig.1.1}
\includegraphics[width=0.225\textwidth]{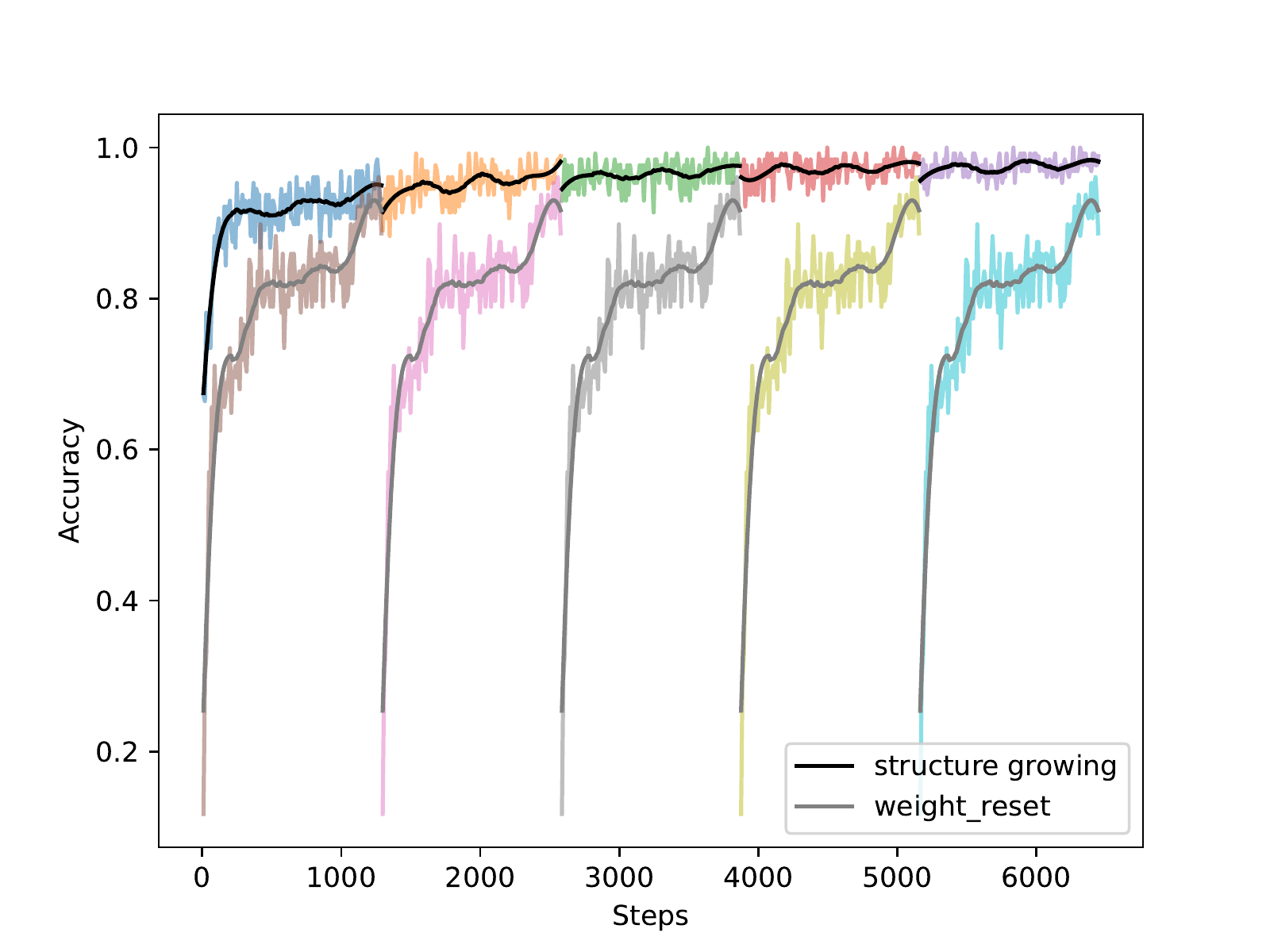}}
\subfigure[validating]{
\label{Fig.1.2}
\includegraphics[width=0.225\textwidth]{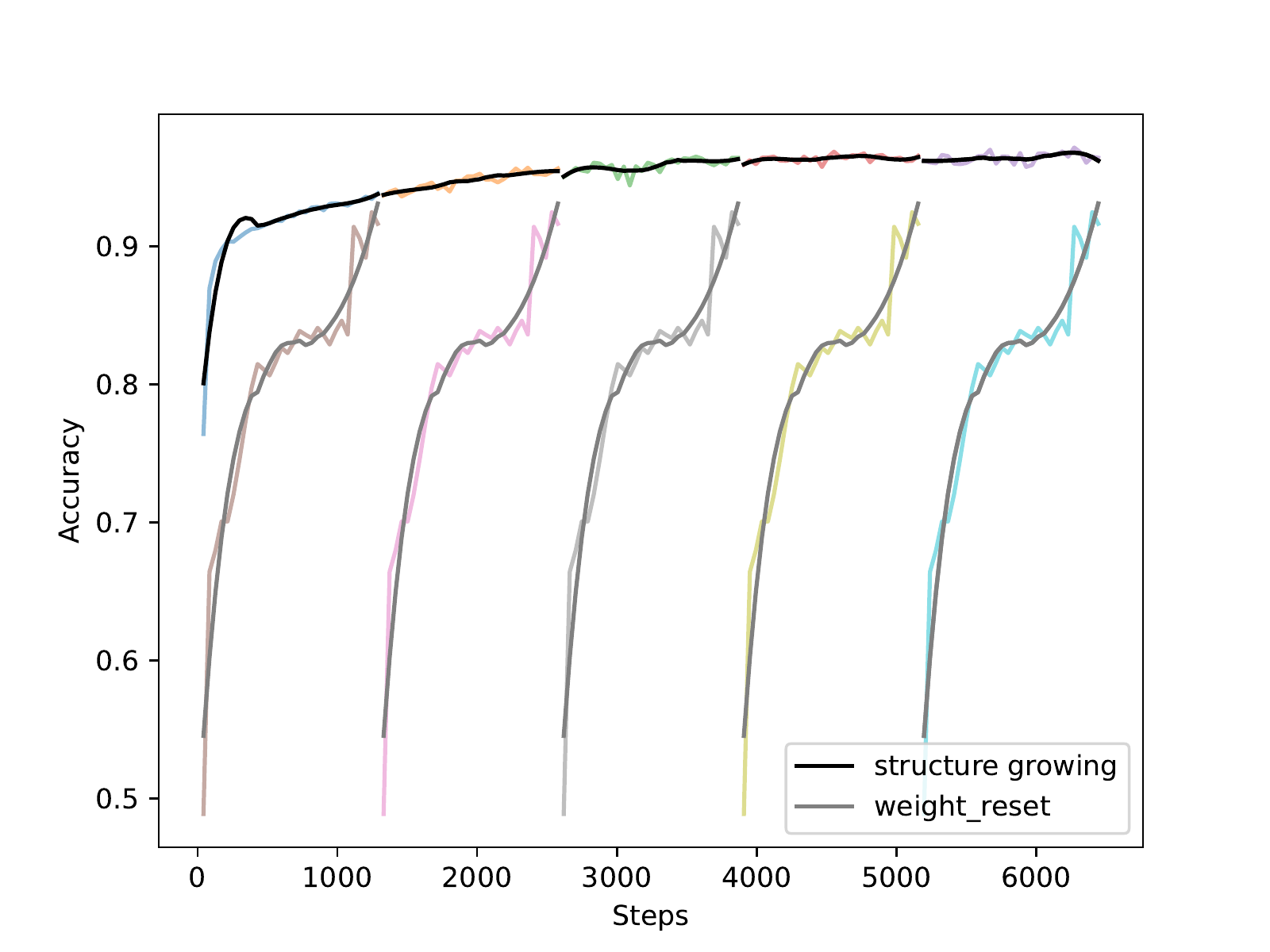}}
\caption{Accuracy on MNIST Dataset}
\label{Fig.1}
\end{figure}

\begin{figure}[htbb]
\subfigure[training]{
\label{Fig.2.1}
\includegraphics[width=0.225\textwidth]{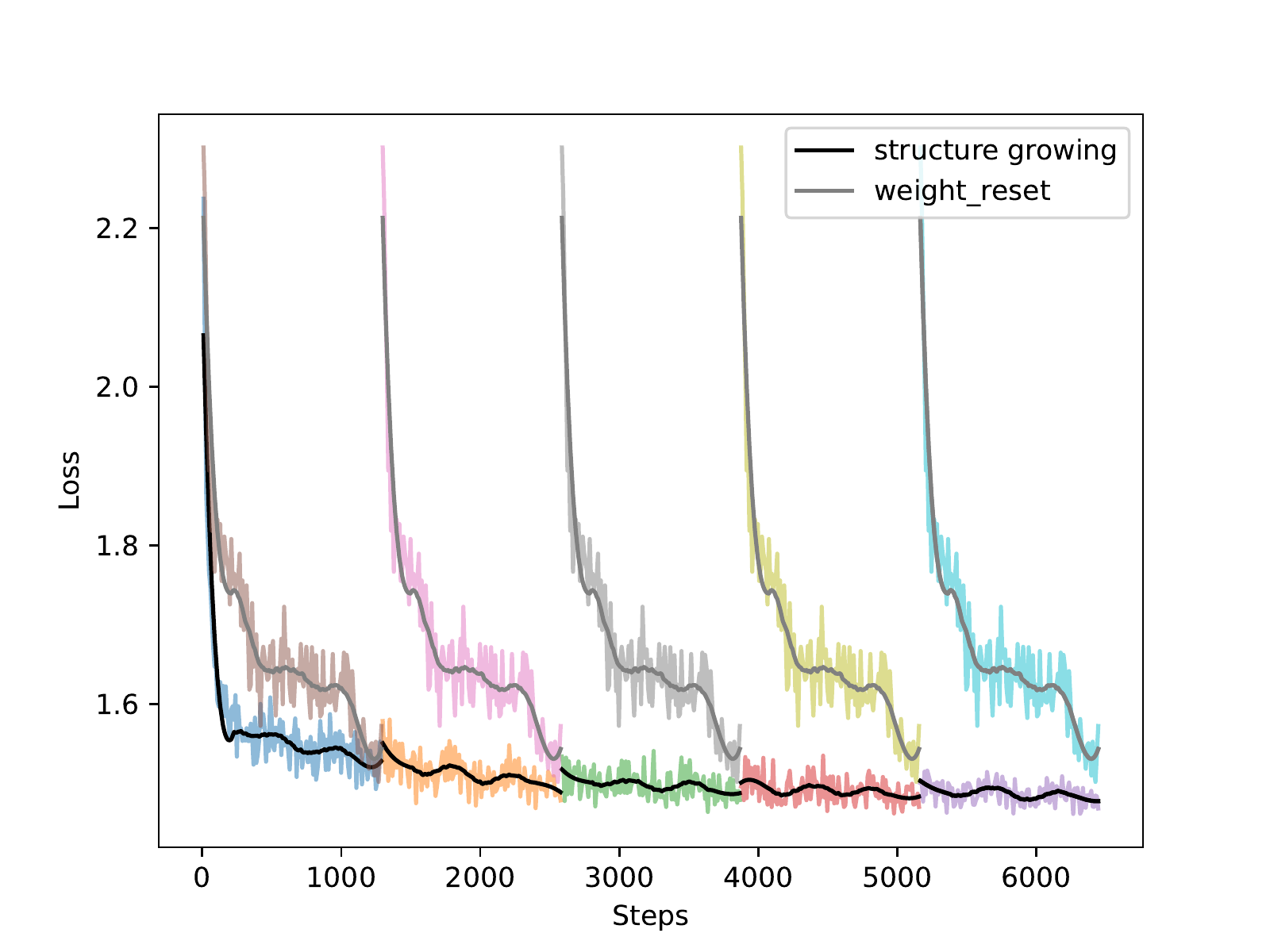}}
\subfigure[validating]{
\label{Fig.2.2}
\includegraphics[width=0.225\textwidth]{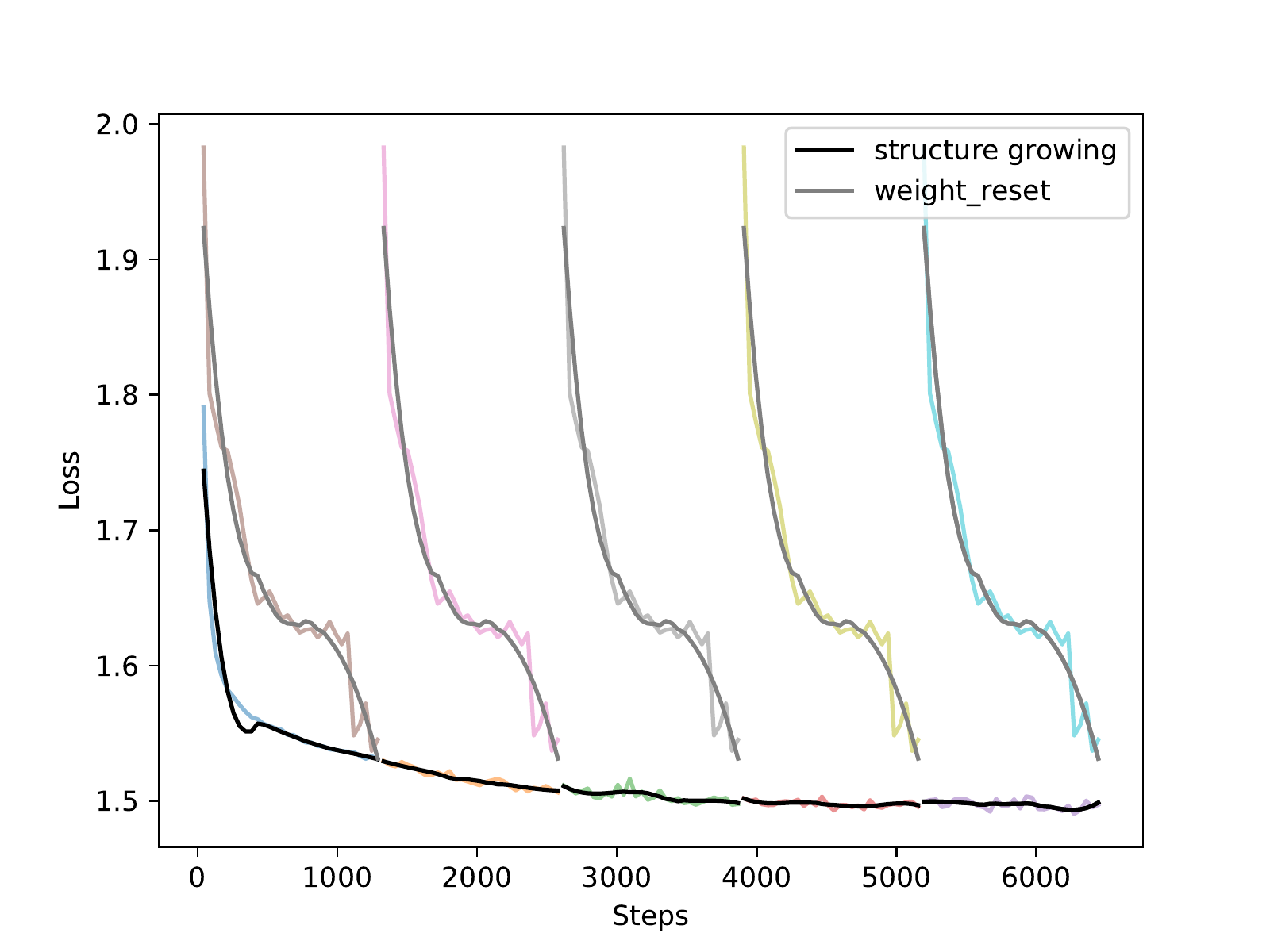}}
\caption{Loss on MNIST Dataset}
\label{Fig.2}
\end{figure}

The loss and accuracy changes during training are shown in the Figure \ref{Fig.1} and Figure \ref{Fig.2}, different color referring to different training iterations.

\subsection{RLTH on ResNet and Convolution Networks}
This part of experiment was executed on CIFAR-10 Dataset \citep{krizhevsky2009learning}. To fitting the multi-channel image data of this dataset. The input and output layer was different form above. And training process was iteratively executed for 7 times.

The neural network to be trained with weight reset policy iteratively is a random-initialized residual convolution neural network. The network have a input layer with input channel of 3 and output channel of 8, and kernel size is 7. Thus those 7 hidden layers have been set to basicblocks of ResNet \citep{he_deep_2016} with input channel of 8 and output channel of 8. Finally, the output layer produce the category probability of the output image with a Softmax activation layer.

And The neural network to be trained with structure growing policy have the same input and output layer and have no hidden layers at begin. When iteratively training, hidden layers would be insert to the network via so-called structure growing operation, too. We insert a basicblocks of ResNet with input channel of 8 and output channel of 8 per iteration, the convolution kernels of inserted layer was initialized with a identity matrix for weight.

The other settings of the experiment are the same with above except the sparsity ratio chaged to 80\% and the pruning algorithms only compress hidden layers of convolution Basicblocks.

\begin{figure}[htbb]
\subfigure[training]{
\label{Fig.3.1}
\includegraphics[width=0.225\textwidth]{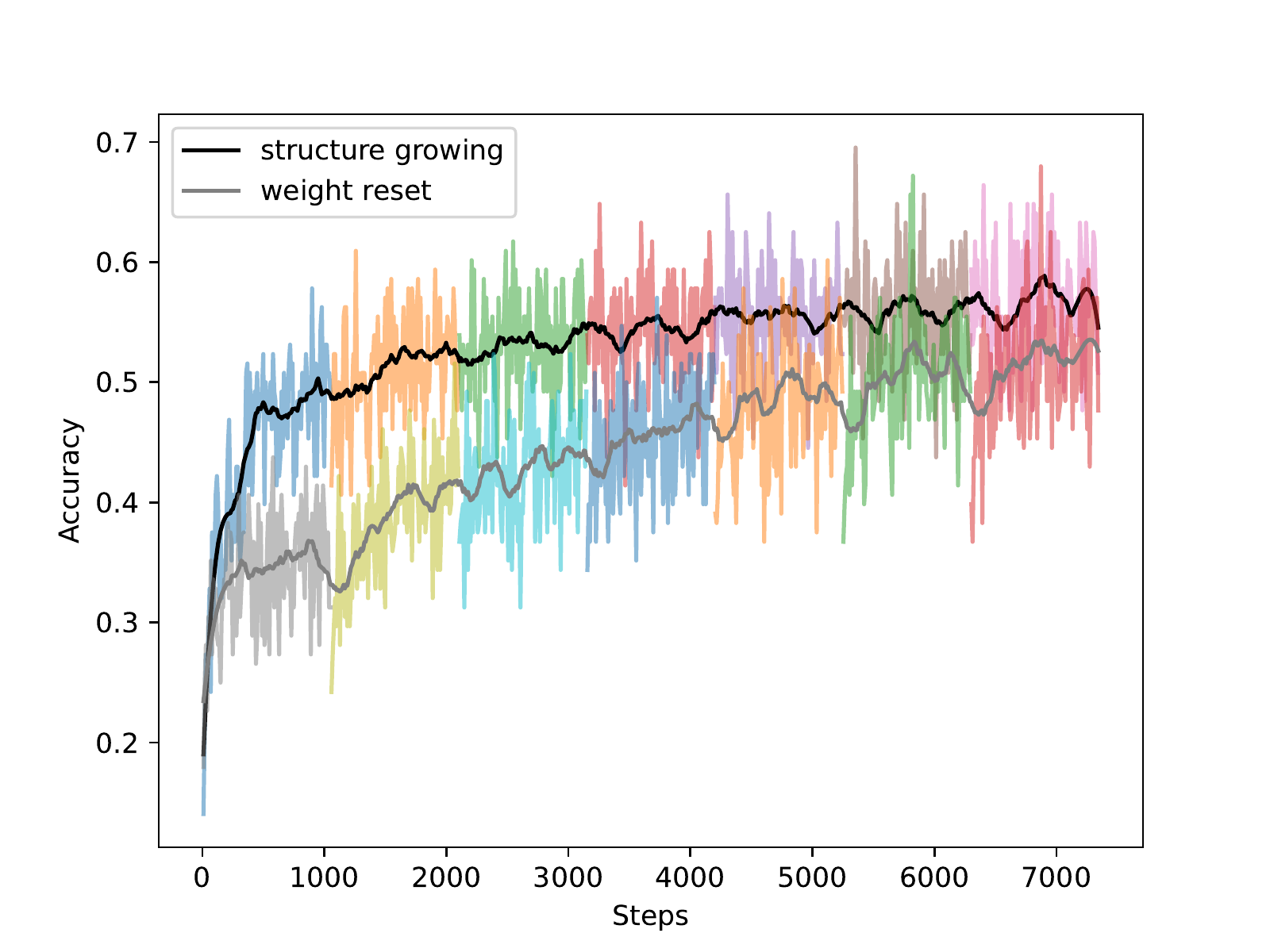}}
\subfigure[validating]{
\label{Fig.3.2}
\includegraphics[width=0.225\textwidth]{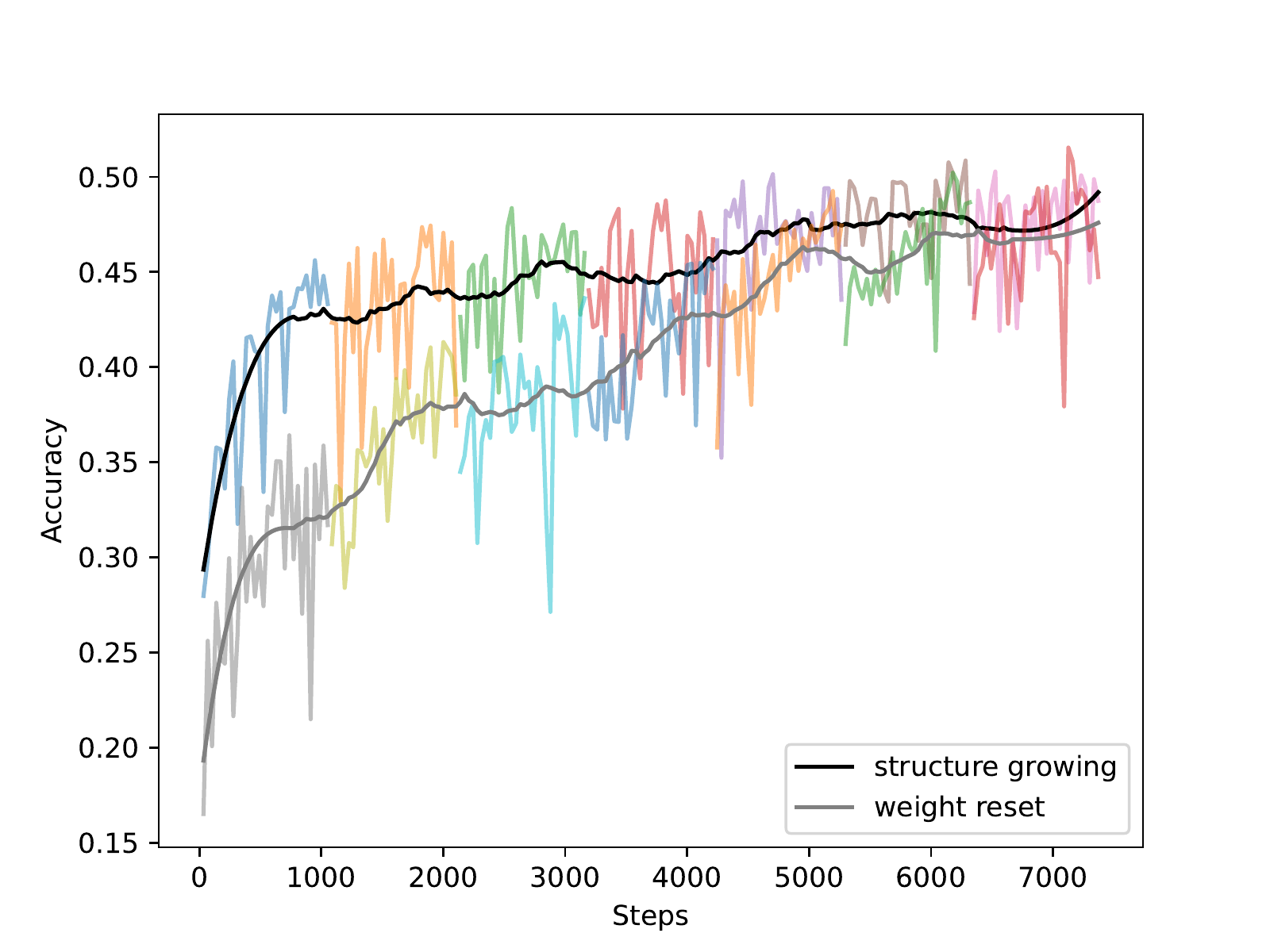}}
\caption{Accuracy on CIFAR-10 Dataset}
\label{Fig.3}
\end{figure}

\begin{figure}[htbb]
\subfigure[training]{
\label{Fig.4.1}
\includegraphics[width=0.225\textwidth]{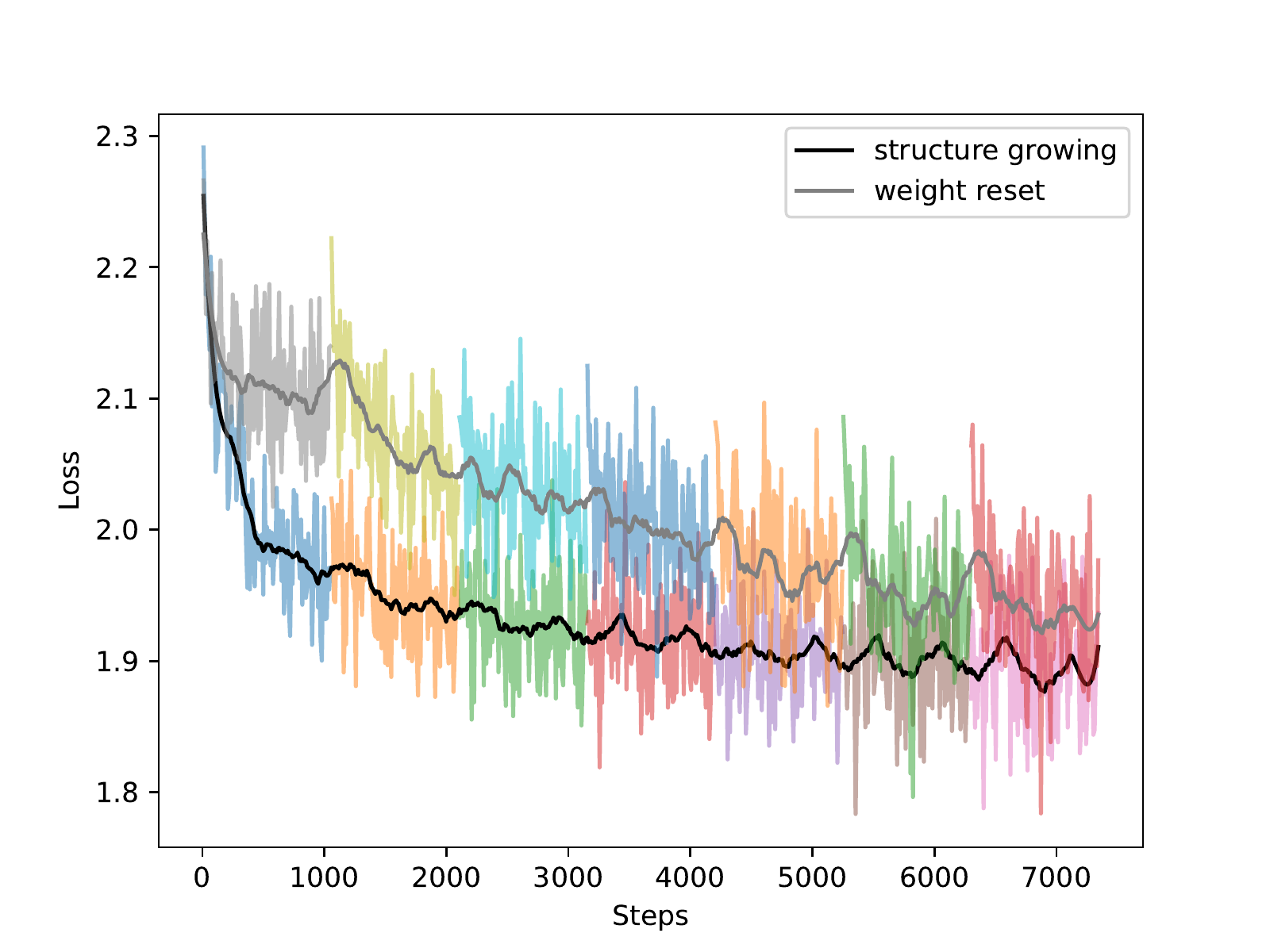}}
\subfigure[validating]{
\label{Fig.4.2}
\includegraphics[width=0.225\textwidth]{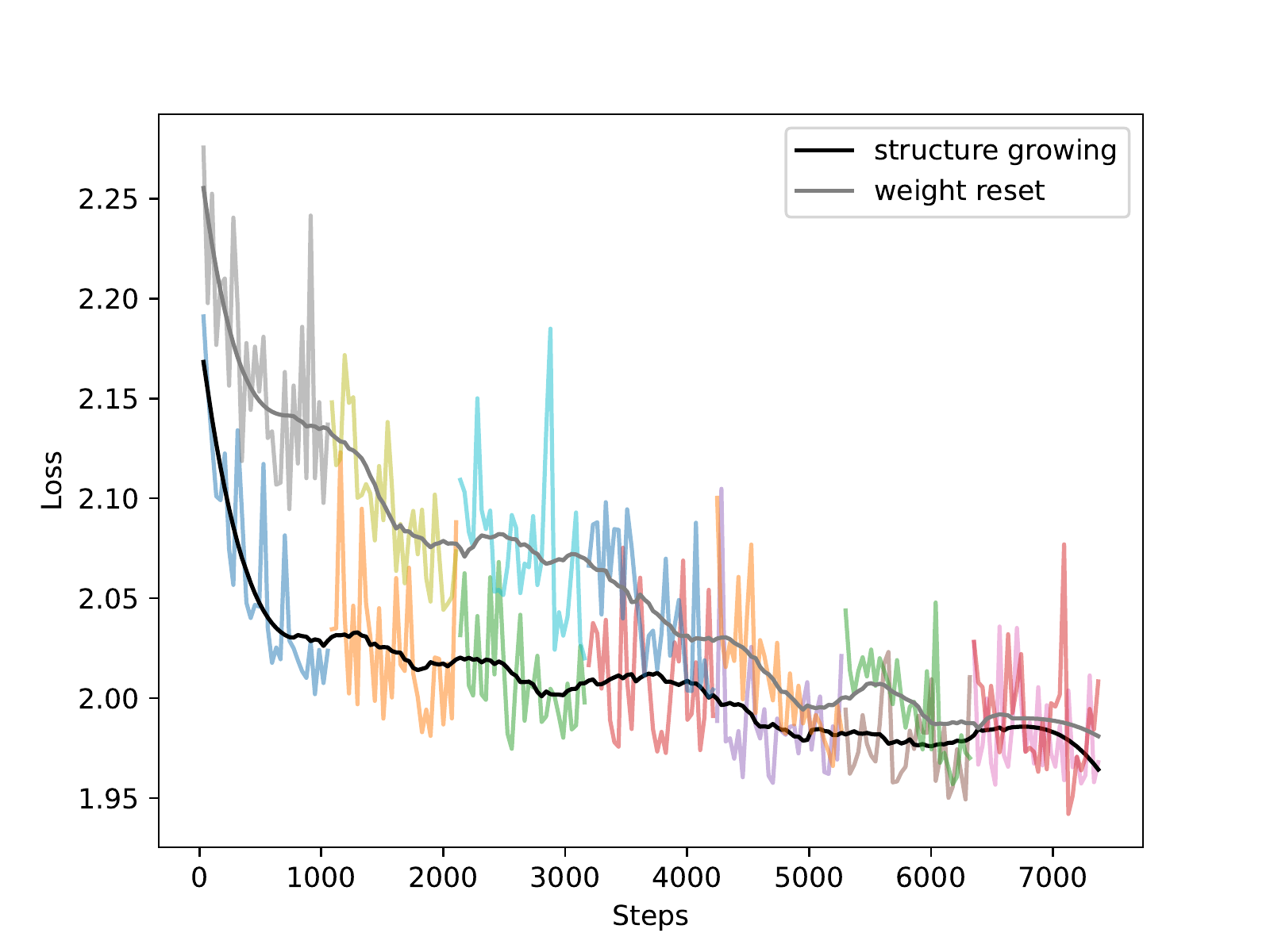}}
\caption{Loss on CIFAR-10 Dataset}
\label{Fig.4}
\end{figure}

The loss and accuracy changes during training are shown in the Figure \ref{Fig.3} and Figure \ref{Fig.4}, different color referring to different training iterations.

From the above two parts of the experiments, we can see that although the weight reset policy avoids the training difficulties of the pruned neural network, its forgetting of knowledge is irrecoverable, and the forgetting slows down the training process and the final test performance when the \emph{winning ticket} sub-networks are repeatedly acquired and trained. In contrast, the structural growth strategy avoids the training difficulties of the pruned neural network by adding new trainable parameters after the \emph{winning ticket} sub-networks are acquired, and ensuring the transfer of the learned knowledge at the same time. This policy is useful for accelerating convergence and reducing overhead in the early training of large neural networks.

\section{Discussion}

\subsection{Juvenile state of neural networks}
Based on the above experiments, we can see that the weight reset policy and the structure growing policy are similar in that both strategies play a similar role in the training of neural networks and the acquisition of \emph{winning ticket}. To clarify this role, we need to go back to the early starting point of frankle's work, i.e., to overcome the training difficulties and test performance degradation faced by the re-training of pruned neural networks. And the weight reset policy address this difficulty by re-empowering the pruned neural networks to obtain the speed of convergence and test performance comparable to the original neural networks just after initialized.

Therefore, we can make a qualitative hypothesis that when the initialization method and network structure are appropriate, the original initialized neural network has a property that has been weakened during the pruning or improper training, and it is this property that affects the so-called learning potential, also known as the ability of the network to converge at a better test accuracy with an appropriate training policy.

As for this property, when applied to the learning process of neural networks, it has a high degree of similarity to the learning process of natural animals, so we can introduce a bio-behavioral phenomena, juvenile state, to define this property, where the learning skills and proficiency of young animals progress significantly faster than those of adults \citep{delacour_memory_1994,boesch_learning_2019}. The main manifestation of this in humans is in language learning, where young children who have not acquired any language skills learn their native language significantly faster than adults who learn the same language as a foreign language \citep{snow_mothers_1972,leung_parents_2021}. For canines, there is a golden period of training, with puppies around half a year old acquiring tricks and skills much faster than adult dogs after sexual maturity \citep{dogyoucanteach,wallis_aging_2016}.

For neural networks, the juvenile state is the state where the model has good learning potential and fast convergence speed. When the neural network is in the juvenile state or the juvenile state has good properties, the neural network can converge at a good test performance level by a fast speed with appropriate learning strategies and optimizers.

In contrast, the juvenile state or property of juvenile of a neural network is weakened during a inappropriate training process or model pruning, which is most intuitively manifested by the difficulty of convergence and the degradation of test performance when the neural network continues training after pruning.
\subsection{winning tickets and juvenile state}
Returning to the lottery ticket hypothesis, we can consider that the \emph{winning ticket} sub-networks are the sub-networks with better learning potential among the different sub-networks obtained from the same original neural network. Thus the weight reset policy and structure growing policy, to some extent, exerted a rejuvenation of the sub-networks by modifying the sub-networks structure or weights to restore the learning potential of the sub-networks that were damaged during the pruning process.

Therefore, based on the above hypothesis and analysis, we can consider that the process of obtaining a \emph{winning ticket} sub-networks is to obtain a sub-networks with smaller structure and smaller parameters amount from a neural network who was already converged. And at the same time, the juvenile properties of this sub-network would also close to the original just initialized neural network. A good \emph{winning ticket} sub-networks has a good juvenile properties to ensure it can converge as fast as the original just initialized neural network with a smaller parameters amount and a comparable test performance after convergence.

\subsection{Factors affecting Juvenile state}
In the above analysis, we have covered some behaviors that are harmful or beneficial to the neural network juvenile state and properties. In the following, we would discuss those factors that may affect the neural network's juvenile state.

First, the weights and initialization of the neural network. The most important contribution of the work of frankle et al. \citep{frankle_lottery_2019} is that the convergence speed and testing performance of the network are highly correlated with the initialization process of the network, and the weight resetting policy restores the property of juvenile and learning potential of the network precisely by resetting the network to the initial weights after pruning.

Second, the structure of the neural network. The impact of the structure of the neural network on the neural network juvenile properties is reflected in the common sense. When a model structure is not suitable for training data, the neural network will face the situation of training difficult or degraded test performance even from the beginning of training.

Thus, If the number of model parameters is too large, it would make the model difficult to train and slow to converge. While if the number of parameters is too small, it would make the neural network easy to fall into the local optima and cause a overfitting of the model.

In contrast, experiments on the structural growth strategy show that appropriate overparameterization of structure is helpful for improving the property of juvenile of the model.

Third, the training strategy. In our usual sense, It contains the loss function, the optimization algorithm and its hyperparameter settings, such as learning rate and momentum, as well as other settings such as batch size and data set sampling methods appeared in deep learning practices.

Specifically, just for the gradient descent method, from a viewpoint of the \emph{loss function surface landscape} \citep{DBLP:journals/corr/abs-1712-09913}, the projection points (or the center of gravity of multiple projection points) of the same (or the same batch) dataset sample points on the loss surface, during the iterative optimization process, are also in constant motion, and ideally most of the projection points can reach a local elevation minimum along the fastest descending line, i.e., the model converges on a local optimal solution. And models can leave the local optima by the momentum methods to reach a better approximate solution.

The proper loss function and proper training policy provided a good \emph{loss function surface landscape}, which enables faster convergence, avoidance of local optima traps and better test performance of the neural network in the above process, which corresponds to the juvenile state assumption.

Finally, better test scores tend to reflect better generalization performance for different \emph{winning ticket} sub-networks within the same dataset. on the other hand, the work of Ibrahim et al. \citep{alabdulmohsin_generalized_2021,morcos2019one} shown the existence of \emph{winning ticket} sub-networks with the ability to generalize across datasets and training policies. So, we can assume that the learning potential of a neural network and juvenile state properties on datasets also affects the neural network's generalization ability.

\section{Conclusion}
In this paper,we introduced the recursive lottery ticket hypothesis by generalizing the lottery ticket hypothesis for neural networks. By recursively acquiring the \emph{winning ticket} sub-networks and continuing the training after add new trainable parameters, we can avoid the training difficulty of pruned network and the knowledge forgetting of the lottery ticket hypothesis does.

From the perspective of connectionism,the recursive lottery ticket hypothesis uses the pruning algorithm as a fine-grained optimizer of the link structure between neuron layers.The structure growth operation allows the introduction of more complex neural architecture search algorithms to above procedure,thus a bridge between structural search and structural optimization have been constructed. Finally, we further propose the juvenile state hypothesis by linking previous lottery ticket hypothesis studies with similar phenomena in biological behavior, and finally we made a qualitative analysis of those factors may affecting the juvenile state of neural networks.

\section{Future works}
Due to hardware and time constraints, the datasets and pruning algorithms involved in the experiments and the architectural search operations in this paper are relatively simple and naive. In the near future, we will verify that hypothesis on more complex problems and introduce more complex algorithms for the structural operations of neural networks.

In addition, another terminology closely related to juvenile state in biology is Neoteny \citep{shea_heterochrony_1989,price_behavioral_1999}, also known as juvenile state's continuation, which is a state in which an organism can maintain its physical and mental young and strong learning ability for a long time even last for life, similar to continuous learning in machine learning that can combat with catastrophic forgettings \citep{9349197}, and it is one of the main directions of our future work.

% \section*{References}

\bibliography{ref.bib}

\begin{thebibliography}{26}
\providecommand{\natexlab}[1]{#1}

\bibitem[{Alabdulmohsin et~al.(2021)Alabdulmohsin, Markeeva, Keysers, and
  Tolstikhin}]{alabdulmohsin_generalized_2021}
Alabdulmohsin, I.; Markeeva, L.; Keysers, D.; and Tolstikhin, I. 2021.
\newblock A {Generalized} {Lottery} {Ticket} {Hypothesis}.
\newblock \emph{arXiv:2107.06825 [cs]}.
\newblock ArXiv: 2107.06825.

\bibitem[{Boesch et~al.(2019)Boesch, Bombjaková, Meier, and
  Mundry}]{boesch_learning_2019}
Boesch, C.; Bombjaková, D.; Meier, A.; and Mundry, R. 2019.
\newblock Learning curves and teaching when acquiring nut-cracking in humans
  and chimpanzees.
\newblock \emph{Scientific Reports}, 9(1): 1515.
\newblock Bandiera\_abtest: a Cc\_license\_type: cc\_by Cg\_type: Nature
  Research Journals Number: 1 Primary\_atype: Research Publisher: Nature
  Publishing Group Subject\_term: Behavioural ecology;Biological anthropology
  Subject\_term\_id: behavioural-ecology;biological-anthropology.

\bibitem[{Cohen and Shashua(2016)}]{cohen_inductive_2016}
Cohen, N.; and Shashua, A. 2016.
\newblock Inductive bias of deep convolutional networks through pooling
  geometry.
\newblock Edition: arXiv preprint arXiv: 1605.06743.

\bibitem[{Delacour(1994)}]{delacour_memory_1994}
Delacour, J. 1994.
\newblock \emph{The {Memory} {System} of the {Brain}}.
\newblock World Scientific.
\newblock ISBN 978-981-02-1021-2.
\newblock Google-Books-ID: CCtqDQAAQBAJ.

\bibitem[{Delange et~al.(2021)Delange, Aljundi, Masana, Parisot, Jia,
  Leonardis, Slabaugh, and Tuytelaars}]{9349197}
Delange, M.; Aljundi, R.; Masana, M.; Parisot, S.; Jia, X.; Leonardis, A.;
  Slabaugh, G.; and Tuytelaars, T. 2021.
\newblock A continual learning survey: Defying forgetting in classification
  tasks.
\newblock \emph{IEEE Transactions on Pattern Analysis and Machine
  Intelligence}, 1--1.

\bibitem[{Frankle and Carbin(2019)}]{frankle_lottery_2019}
Frankle, J.; and Carbin, M. 2019.
\newblock The {Lottery} {Ticket} {Hypothesis}: {Finding} {Sparse}, {Trainable}
  {Neural} {Networks}.
\newblock \emph{arXiv:1803.03635 [cs]}.
\newblock ArXiv: 1803.03635.

\bibitem[{Frankle et~al.(2019)Frankle, Dziugaite, Roy, and
  Carbin}]{frankle2019stabilizing}
Frankle, J.; Dziugaite, G.~K.; Roy, D.~M.; and Carbin, M. 2019.
\newblock Stabilizing the lottery ticket hypothesis.
\newblock \emph{arXiv preprint arXiv:1903.01611}.

\bibitem[{Han et~al.(2015)Han, Pool, Tran, and
  Dally}]{DBLP:journals/corr/HanPTD15}
Han, S.; Pool, J.; Tran, J.; and Dally, W.~J. 2015.
\newblock Learning both Weights and Connections for Efficient Neural Networks.
\newblock \emph{CoRR}, abs/1506.02626.

\bibitem[{He et~al.(2016)He, Zhang, Ren, and Sun}]{he_deep_2016}
He, K.; Zhang, X.; Ren, S.; and Sun, J. 2016.
\newblock Deep residual learning for image recognition.
\newblock In \emph{Proceedings of the {IEEE} conference on computer vision and
  pattern recognition}, 770--778.

\bibitem[{Kingma and Ba(2017)}]{kingma2017adam}
Kingma, D.~P.; and Ba, J. 2017.
\newblock Adam: A Method for Stochastic Optimization.
\newblock arXiv:1412.6980.

\bibitem[{Krizhevsky, Hinton et~al.(2009)}]{krizhevsky2009learning}
Krizhevsky, A.; Hinton, G.; et~al. 2009.
\newblock Learning multiple layers of features from tiny images.

\bibitem[{Lecun et~al.(1998)Lecun, Bottou, Bengio, and
  Haffner}]{lecun_gradient-based_1998}
Lecun, Y.; Bottou, L.; Bengio, Y.; and Haffner, P. 1998.
\newblock Gradient-based learning applied to document recognition.
\newblock \emph{Proceedings of the IEEE}, 86(11): 2278--2324.

\bibitem[{LeCun, Denker, and Solla(1990)}]{lecun_optimal_1990}
LeCun, Y.; Denker, J.~S.; and Solla, S.~A. 1990.
\newblock Optimal brain damage.
\newblock In \emph{Advances in neural information processing systems},
  598--605.

\bibitem[{Leung, Tunkel, and Yurovsky(2021)}]{leung_parents_2021}
Leung, A.; Tunkel, A.; and Yurovsky, D. 2021.
\newblock Parents {Fine}-{Tune} {Their} {Speech} to {Children}’s {Vocabulary}
  {Knowledge}.
\newblock \emph{Psychological Science}, 32(7): 975--984.
\newblock Publisher: SAGE Publications Inc.

\bibitem[{Li et~al.(2016)Li, Kadav, Durdanovic, Samet, and
  Graf}]{DBLP:journals/corr/LiKDSG16}
Li, H.; Kadav, A.; Durdanovic, I.; Samet, H.; and Graf, H.~P. 2016.
\newblock Pruning Filters for Efficient ConvNets.
\newblock \emph{CoRR}, abs/1608.08710.

\bibitem[{Li et~al.(2017)Li, Xu, Taylor, and
  Goldstein}]{DBLP:journals/corr/abs-1712-09913}
Li, H.; Xu, Z.; Taylor, G.; and Goldstein, T. 2017.
\newblock Visualizing the Loss Landscape of Neural Nets.
\newblock \emph{CoRR}, abs/1712.09913.

\bibitem[{Liang et~al.(2021)Liang, Glossner, Wang, Shi, and
  Zhang}]{LIANG2021370}
Liang, T.; Glossner, J.; Wang, L.; Shi, S.; and Zhang, X. 2021.
\newblock Pruning and quantization for deep neural network acceleration: A
  survey.
\newblock \emph{Neurocomputing}, 461: 370--403.

\bibitem[{Microsoft(2021)}]{noauthor_microsoftnni_2021}
Microsoft. 2021.
\newblock microsoft/nni.
\newblock Original-date: 2018-06-01T05:51:44Z.

\bibitem[{Morcos et~al.(2019)Morcos, Yu, Paganini, and Tian}]{morcos2019one}
Morcos, A.~S.; Yu, H.; Paganini, M.; and Tian, Y. 2019.
\newblock One ticket to win them all: generalizing lottery ticket
  initializations across datasets and optimizers.
\newblock \emph{arXiv preprint arXiv:1906.02773}.

\bibitem[{of~Veterinary Medicine~Vienna(2016)}]{dogyoucanteach}
of~Veterinary Medicine~Vienna, U. 2016.
\newblock You can teach an old dog new tricks, but younger dogs learn faster.

\bibitem[{Price(1999)}]{price_behavioral_1999}
Price, E.~O. 1999.
\newblock Behavioral development in animals undergoing domestication.
\newblock \emph{Applied Animal Behaviour Science}, 65(3): 245--271.

\bibitem[{Ren et~al.(2021)Ren, Xiao, Chang, Huang, Li, Chen, and
  Wang}]{ren2021comprehensive}
Ren, P.; Xiao, Y.; Chang, X.; Huang, P.-Y.; Li, Z.; Chen, X.; and Wang, X.
  2021.
\newblock A comprehensive survey of neural architecture search: Challenges and
  solutions.
\newblock \emph{ACM Computing Surveys (CSUR)}, 54(4): 1--34.

\bibitem[{Renda, Frankle, and Carbin(2020)}]{renda2020comparing}
Renda, A.; Frankle, J.; and Carbin, M. 2020.
\newblock Comparing rewinding and fine-tuning in neural network pruning.
\newblock \emph{arXiv preprint arXiv:2003.02389}.

\bibitem[{Shea(1989)}]{shea_heterochrony_1989}
Shea, B.~T. 1989.
\newblock Heterochrony in human evolution: {The} case for neoteny reconsidered.
\newblock \emph{American Journal of Physical Anthropology}, 32(S10): 69--101.
\newblock \_eprint:
  https://onlinelibrary.wiley.com/doi/pdf/10.1002/ajpa.1330320505.

\bibitem[{Snow(1972)}]{snow_mothers_1972}
Snow, C.~E. 1972.
\newblock Mothers' {Speech} to {Children} {Learning} {Language}.
\newblock \emph{Child Development}, 43(2): 549--565.
\newblock Publisher: [Wiley, Society for Research in Child Development].

\bibitem[{Wallis et~al.(2016)Wallis, Virányi, Müller, Serisier, Huber, and
  Range}]{wallis_aging_2016}
Wallis, L.~J.; Virányi, Z.; Müller, C.~A.; Serisier, S.; Huber, L.; and
  Range, F. 2016.
\newblock Aging effects on discrimination learning, logical reasoning and
  memory in pet dogs.
\newblock \emph{AGE}, 38(1): 6.

\end{thebibliography}
\end{document}